\pdfoutput=1

\documentclass[11pt]{article}
\usepackage[]{ACL2024}

\usepackage{float}
\usepackage{times}
\usepackage{latexsym}
\usepackage[T1]{fontenc}
\usepackage[utf8]{inputenc}
\usepackage{microtype}
\usepackage{inconsolata}
\usepackage{booktabs}
\usepackage{multirow}
\usepackage{tabularx}
\usepackage{xcolor}
\usepackage{array}
\usepackage{xspace}
\usepackage{graphicx}
\usepackage{pgfplots}
\usepackage{pdfpages}
\pgfplotsset{compat=1.9,width=3cm}
\usepackage{subcaption}
\usepackage{enumitem}
\usepackage{amsfonts}
\usepackage{amsmath}
\usepackage{hyperref}
\usepackage{multirow}
\usepackage[export]{adjustbox}
\usepackage{blindtext}
\usepackage{makecell}

\newenvironment{boxed2*}
    {\begin{center}
    \begin{tabular}{|p{0.475\textwidth}|}
    \hline\\
    }
    { 
    \\\\\hline
    \end{tabular} 
    \end{center}
    }

\title{Chart-based Reasoning: Transferring Capabilities from LLMs to VLMs}

\author{Victor Cărbune\thanks{\;\;Correspondence to: \href{mailto:vcarbune@google.com}{vcarbune@google.com}}
\qquad Hassan Mansoor \qquad Fangyu Liu \qquad Rahul Aralikatte\\
\textbf{Gilles Baechler \qquad Jindong Chen \qquad Abhanshu Sharma} \\
Google Research
}

\begin{document}

\maketitle
\begin{abstract}%

Vision-language models (VLMs) are achieving increasingly strong performance on multimodal tasks. However, reasoning capabilities remain limited particularly for smaller VLMs, while those of large-language models (LLMs) have seen numerous improvements. We propose a technique to transfer capabilities from LLMs to VLMs.
On the recently introduced ChartQA, our method obtains state-of-the-art performance when applied on the PaLI3-5B VLM by \citet{chen2023pali3}, while also enabling much better performance on PlotQA and FigureQA.

We first improve the chart representation by continuing the pre-training stage using an improved version of the chart-to-table translation task by \citet{liu2023deplot}. We then propose constructing a 20x larger dataset than the original training set. To improve general reasoning capabilities and improve numerical operations, we synthesize reasoning traces using the table representation of charts. Lastly, our model is fine-tuned using the multitask loss introduced by \citet{hsieh2023distilling}.

Our variant ChartPaLI-5B outperforms even 10x larger models such as PaLIX-55B without using an upstream OCR system, while keeping inference time constant compared to the PaLI3-5B baseline. When rationales are further refined with a simple program-of-thought prompt \cite{chen2023program}, our model outperforms the recently introduced Gemini Ultra and GPT-4V.

\end{abstract}

\section{Introduction}

Visual language, where text and images work together to deliver information, can be expressed through charts, plots, and diagrams. Multimodal reasoning within this context is challenging, as it involves linking visual properties (like color, line style, and positioning) with textual content (such as legends and units).

Many recent advances of vision-language models (VLMs) come from techniques enabling better representations \cite{dosovitskiy2021image, lee2023pix2struct}, giving the model the ability to understand core elements of the image, a necessary building block for basic reasoning. However, complex reasoning capabilities which combine the core representation of the image with semantic understanding of a question to provide an answer, have been rather limited. Models oftentimes are not able to contextually combine image and text representations. One technique that improves reasoning capabilities in large-language models (LLMs) includes in-context learning for eliciting reasoning such as chain-of-thought prompting \cite{wei2023chainofthought}, decomposing tasks \cite{zhou2023leasttomost} or composing stored facts in weights \cite{press2023measuring}. Fine-tuning on datasets with rationales \cite{magister2023teaching, hsieh2023distilling} has been shown to be effective for smaller models. In this work, we tackle improving reasoning capabilities in VLMs through better learned image representations, followed by fine-tuning on synthetic datasets with reasoning traces generated by more capable LLMs. We also explore a hybrid online setup for numerical reasoning refinements. 

\begin{figure}
\includegraphics[width=\columnwidth]{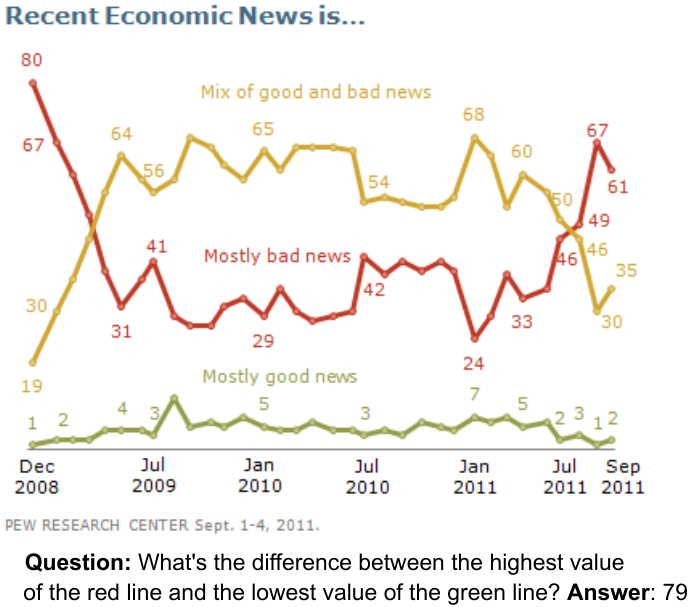}
\caption{Example from the ChartQA validation set.}
\label{fig:chartqa_example}
\end{figure}

\noindent We empirically show that this indeed improves performance through experiments on ChartQA \cite{masry2022chartqa}. Visual-question answering on charts quantifies the ability of a VLM to reason using complex information presented. Oftentimes answering the question requires implicit or explicit information extraction, followed by intermediate grouping or computations using the extracted information, and reasoning with the final quantities, as shown in Figure~\ref{fig:chartqa_example}.

\noindent Vision-language models (VLMs) such as PaLI-X and PaLI-3 are hybrid model architectures which use a vision and a language backbone to solve visual tasks \cite{chen2023palix, chen2023pali3}. The training recipe typically involves a pre-training stage focused on learning a good internal representation, followed by a downstream fine-tuning stage. \citet{chen2023pali3} note that PaLI-3 falls behind PaLI-X on ChartQA likely due to its limited reasoning capabilities. Results presented in this work suggest that the lack of a pre-training task for learning better chart representations, as done in \citet{liu2023matcha}, may be another reason.

\noindent Enhancing the reasoning capabilities of large language models (LLMs) such as PaLM-2 \cite{palm2} or GPT-4 \cite{openai2023gpt4} is a very active research area. While reasoning is considered an emerging property with scale \cite{wei2022emergent}, \citet{press2023measuring} argue that simply scaling only enables better memorization of knowledge and does not enable composing multiple stored facts into an answer. On the other hand, prompting techniques enacting complex reasoning on downstream tasks have been shown to be very effective  \cite{wei2023chainofthought} \cite{zhou2023leasttomost}.

Transferring reasoning capabilities from large to small models enables reducing serving costs, while increasing task performance. \citet{hsieh2023distilling} have introduced an effective multi-task framework which enable small models to outperform their much larger counterparts using less data. They do so by leveraging rationale generation as a separate task, instead of more standard distillation approaches, which first infer the rationale, followed by the answer \cite{magister2023teaching}. We apply this framework for the first time on multimodal tasks.

\paragraph{Contributions} Our main results can be summarized as follows: \textbf{(i)} we introduce an efficient recipe consisting of a pre-training task and fine-tuning task with synthetic datasets using a multi-task setup for improving reasoning capabilities, \textbf{(ii)} we obtain SoTA performance by significantly improving PaLI-3 performance on the ChartQA benchmark with our recipe and using 10x less parameters than prior work, \textbf{(iii)} we perform numerous ablation experiments quantifying the impact of the techniques used in our recipe.

The remainder of this paper is structured as follows.
Section~\ref{sec:related} describes related work, followed by Section~\ref{sec:dataset} which introduces the construction of the training datasets. Section~\ref{sec:method} illustrates our novel pre-training and fine-tuning recipe, followed by Section~\ref{sec:experiments} describing the experimental setup and main results. Lastly, Section~\ref{sec:conclusion} delivers a conclusion and recommendation for future work, followed by Section~\ref{sec:limitations} where we acknowledge limitations of the current work.

\section{Related Work}\label{sec:related}

\paragraph{VLM landscape} Vision-language models usually combine a vision backbone with a language backbone. Frequently it is a Vision Transformer (ViT) \cite{dosovitskiy2021image} coupled with a Large Language Model via an encoder-decoder \cite{chen2023palix} or decoder-only \cite{alayrac2022flamingo} architecture. More recently, models such as Fuyu-8B \cite{fuyu-8b} explore projecting the image directly through the language backbone. In this work we extend PaLI-3, an encoder-decoder architecture with ViT-3B as vision and UL2-2B as language backbones. We refer the reader to \citet{chen2023pali3} for a complete overview. PaLI-3 is a SoTA model and hence we decided to build on top of it to further focus on improving the results with our methods.

\paragraph{Existing approaches for chart understanding} The task of answering questions on charts is, alongside documents and infographics, part of a broader set of tasks commonly referred to \textit{visually-situated language understanding}, where text and image cannot be treated separately \cite{lee2023pix2struct}. Fine-tuned models on downstream ChartQA include PaLI-3 \cite{chen2023pali3}, MatCha \cite{liu2023matcha} and UniChart \cite{masry2023unichart}. Among these, UniChart takes the most similar approach to ours, pre-training a chart image encoder as vision backbone and BART decoder \cite{lewis2019bart} as language backbone. Alternatively, \citet{liu2023deplot} took the approach of decomposing question-answering into first translating the chart into a table, then querying an LLM in a plug-and-play fashion. Here our main focus is on fine-tuned self-contained models, however we show that a simple refinement using a much larger LLM, continues to improve performance as well.

\paragraph{The role of upstream OCR systems} 
A chart usually has an underlying equivalent tabular representation of the data. However, decoding the tabular representation remains a challenging problem. Alternatively, charts can be passed through an OCR system to extract an unstructured text representation of the image. \cite{Luo2021ChartOCRDE} combine chart-specific extraction logic with an OCR system to extract key information from the charts. As intuitively expected, usually the use of an OCR system improves downstream quality. In this work, we assume the model only has access to the chart image.

\paragraph{Improving chart reasoning with synthetic data} Having the pre-training mixture specialize on chart tasks is effective \cite{liu2023matcha}. We further extend the \textit{chart derendering} task, which translates charts to code or to table. Similar to our approach, \citet{methani2020plotqa} and \citet{masry2023unichart} have made use of programmatic templates to a synthesize complex QA pairs. However, instead of using an LLM to generate chart summaries as in \citet{masry2023unichart}, here we use it to generate additional QA pairs with rationales. These generated examples together with synthetic programmatic examples are key in the pre-training and fine-tune stages of our model.

\section{Dataset} \label{sec:dataset}

\subsection{Brief description of ChartQA}
ChartQA is one of the widely adopted visual question-answering benchmarks for reasoning capabilities of VLMs.

The standard ChartQA benchmark has two components: (a) human set and (b) augmented generated set. The augmented set has been machine generated and is more simplistic in nature than the human set.

The charts in the dataset come from four sources (Statista, Pew, Our World in Data and OECD). Gold tables are available for all sources, except for Pew, where the tables are inferred with ChartOCR model \cite{Luo2021ChartOCRDE}. 
Although we observed mistakes in inferred tables, our method seems to be fairly resilient to them. 

\subsection{Synthetic Generation Methods}

In this work, we use LLMs to synthesize additional examples paired with rationales generated using chain-of-thought prompting. We use the tabular representation of charts present in the training set as a way to mediate the lack of vision input into LLMs. 

The data we synthesize increases the diversity of the original training set, especially with examples that require extracting multiple quantities from the chart and perform reasoning using them.

We combine two approaches that focus on this type of examples, specifically we use a LLM for synthesizing \textit{rationale generation} and \textit{extra question answer} pairs. We also use a programmatic  approach for generating \textit{arithmetic} question answer pairs.

\paragraph{Rationale Generation}

We augment the original training set with synthetic explanations on why an answer is reached. 
We achieve this by using PaLM 2-S to predict a $\mathbf{rationale}$ on an input tuple of $\mathbf{(table, question, answer)}$ with a 4-shot prompt, as illustrated in  Figure~\ref{fig:rationale_prompt}.
We refer to this set as \textit{ChartQA-Rationale-S}. 

\begin{figure}
\includegraphics[width=\columnwidth]{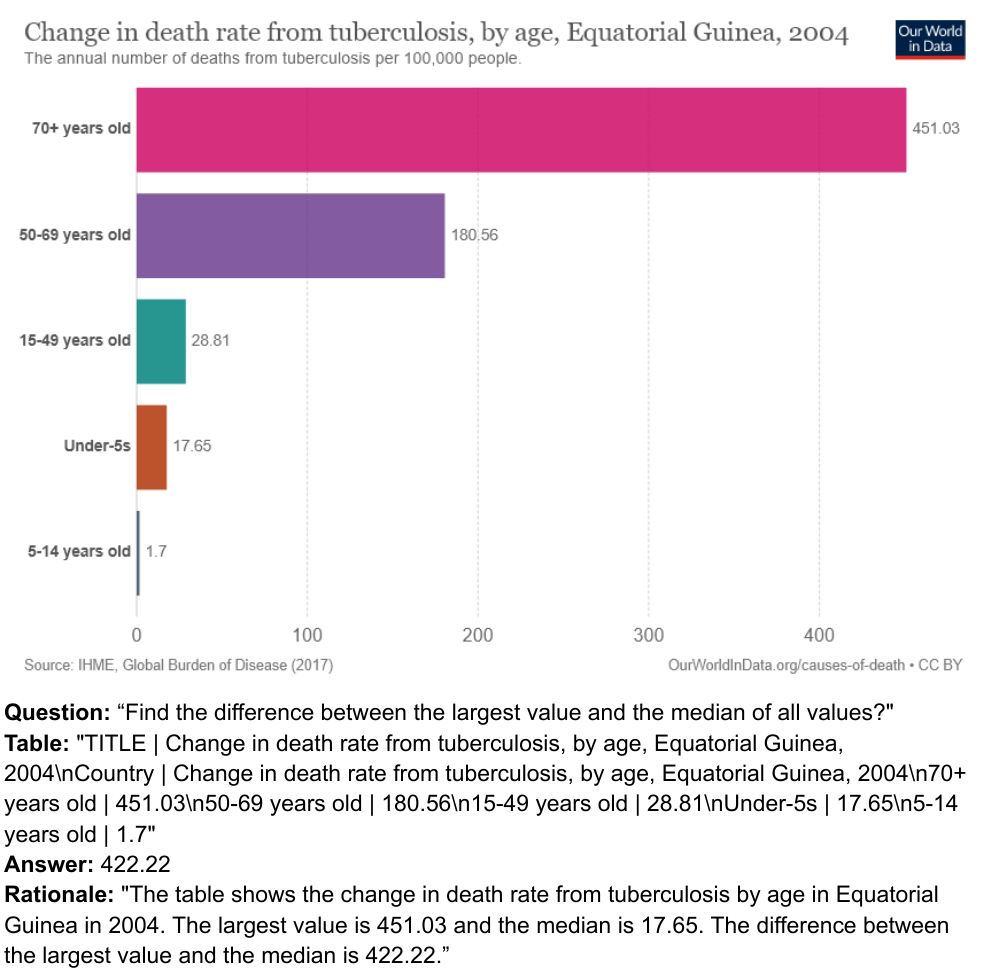}
\caption{\textit{ChartQA-Rationale-S}:
For each example of the original training set, we synthesize a rational based on the table, the question and the answer.
}
\label{fig:rationale_example}
\end{figure}

By requesting the model to provide justifications for ground truth answers, which are typically accurate, we witness a significant reduction in hallucinations.
A notable exception is when the answer itself is wrong, which happens more frequently for the ChartQA augmented set than the human set. However, we did not perform a detailed investigation of this aspect in the generated training sets. An instance of the generated rationale can be seen in Figure~\ref{fig:rationale_example}.

\paragraph{ExtraQA Generation}

We hypothesize that the original training set is too small to contain enough diversity in the examples to enable solving more complex QA questions such as the ones present in the human validation set. Therefore we used a 1-shot prompt illustrated in Figure~\ref{fig:extraqa_prompt} to generate additional examples covering types of errors we identify by examining the model performance on the validation set. 
The prompt is adapted from the one used in \cite{liu2023deplot}. An example of a generated sample can be seen in Figure~\ref{fig:extraqar_example}.
We used both PaLM 2-S and PaLM 2-L to generate the examples and refer to the respective datasets as \textit{ChartQA-ExtraQAR-S/L}.
We perform only lightweight filtering of generated examples that deviate from the imposed structure. If we cannot parse from the LLM response all three elements, we simply drop the example. However, we do not verify the generated examples for hallucinations, fluency or perform any other model-based verification.

\paragraph{ArithmeticQA Generation}

It is well known that large language models have difficulties in performing arithmetic computations accurately. For ChartQA, this is particularly exacerbated by the fact that the small training dataset is adequate for the specifics of the arithmetic questions one can have for charts (as represented by the test set). We programmatically create examples which either require numeric reasoning or a comparative analysis of multiple chart elements. Examples are illustrated in Figure~\ref{fig:arithm_example_1} and Figure~\ref{fig:arithm_example_2}. We abstracted the questions into templates and used a fixed set of mathematical operations such as median, max, min etc. For each template we created a rationale to teach the model a plan to solve the arithmetic problems. For example, computing the mean requires first looking up the values, then adding them up and finally dividing the value by the total.
For each type of arithmetic we created multiple templates both for the questions and rationales. The source data we used are only the ChartQA human examples, using the available tables. The type of questions and their count can be found in Table~\ref{tab:arithmeticqa_split}.

\begin{table}[H]
\centering
\small
\begin{tabular}{lr}
\toprule
\textbf{Question Type} & \textbf{Count} \# \\
\midrule
Mean & 235K \\
Subtraction & 90K \\
Other & 32K \\
\midrule
Total & 357K \\
\bottomrule
\end{tabular}%
\caption{\label{tab:arithmeticqa_split}
Examples are mostly means or subtractions.
}
\end{table}

\subsection{Resulting Dataset}

The resulting dataset is roughly 20x larger and is described in Table~\ref{tab:reasoning_datasets}, with further details on the statistics of the dataset in Section~\ref{sec:statistics_dataset}. Sampling was done using greedy decoding with temperature $\tau=0$. We used the augmented and human sets to generate examples.

\begin{table*}[t]
\centering
\begin{tabular}{lrrlrr}
\toprule
\textbf{Dataset} & \textbf{Hum} \# & \textbf{Aug} \# & \textbf{Question type} \# & \textbf{Total} & \textbf{Rate} \# \\
\midrule
ChartQA-Rationale-S & 7398 & 20901 & {\small R [13\%], V [11\%], C [43\%], B [33\%]} & 28.3K & 15\% \\
ChartQA-ExtraQAR-S & 23261 & 69433 & {\small R [57\%], C [43\%]} & 92.7K & 15\% \\
ChartQA-ExtraQAR-L & 16388 & 50468 & {\small R [60\%], C [40\%]} & 66.9K & 30\% \\
ChartQA-ArithmQAR & 357000 & - & {\small C [100\%]} & 357.0K & 40\% \\
\midrule
ChartQA-Synth (Total) & & & & \textbf{544.9K} & \\
\bottomrule
\end{tabular}
\caption{\label{tab:reasoning_datasets}
Overview of the synthetic dataset, which is 20x larger than the original one. The suffix denotes the size of the PaLM 2 model used. The rate refers to the final mixture. Categorization of question types are from \cite{masry2022chartqa}, namely \textbf{R}etrieval, \textbf{V}isual, \textbf{C}ompositional or \textbf{B}oth visual and compositional.
}
\end{table*}

\paragraph{PaLM 2-S vs. 2-L}
The same prompt was used for all examples in the synthetic dataset. We note that using samples from both LLMs improves performance, but ablation studies do not indicate one is better than the other.  We hypothesize that diversity matters more than model size, but we have not investigated sampling strategies.

\section{Method}\label{sec:method}

Our work builds on top of PaLI-3 architecture and pre-training recipe, which consists of two backbones, a Vision Transformer ViT-2B and Text Encoder-Decoder UL2-3B. Our starting point is the recipe described by \citet{chen2023pali3}. The uni-modal pre-training stage trains the vision encoder using contrastive loss through the SigLIP loss, while the language encoder-decoder is pre-trained using the UL2 loss. Both backbones are pre-trained jointly using a multi-modal stage. Lastly the resolution increase stage enables the vision encoder backbone to work with 812x812 resolution images. We continue pre-training using this checkpoint.

\subsection{Pre-training: Chart2Table Mixture}

Extending the work done by \citet{liu2023deplot}, we use a chart-to-table dataset mixture to continue pre-training with the ViT backbone unfrozen, which facilitates learning an internal representation of the chart. We do not explicitly use the tabular conversion further downstream.

\paragraph{Dataset} For learning this representation, we combine several chart-to-table derendering tasks into a mixture: (1) synthetic chart-to-table data similar to the synthetic mixture introduced by \citet{liu2023deplot}. We traverse different combinations of plotting options in matplotlib and seaborn to randomly plot tables from Wikipedia into charts of different layouts. (2) the chart-to-table mixture introduced by \citet{masry2023unichart}. (3) The chart-table pairs from the train set of DVQA \cite{kafle2018dvqa}. (4) The chart-table pairs from the train set of TaTA \cite{gehrmann2022tata}. (5) The chart-table pairs introduced in Benetech - Making Chart Accessible Kaggle challenge\footnote{\url{https://www.kaggle.com/competitions/benetech-making-graphs-accessible}}.
A complete listing of data source, sampling weight, and number of examples is shown in Table~\ref{tab:chart_to_table_datasets}.

\begin{table}[H]
\centering
\small
\begin{tabular}{lrr}
\toprule
\textbf{Component} & \textbf{Rate} & \textbf{Size} \\
\midrule
Synthetic & 44.0\% & 1.2M \\
UniChart & 39.5\% & 612K \\
DVQA & 3.2\% & 200K \\
ChartQA & 3.2\% & 22K \\
TaTa & 3.2\% & 6.7K \\
Chart2Text & 3.2\% & 24K \\
Benetech Challenge & 3.2\% & 21K \\
PlotQA & 0.5\% & 224K \\
\midrule
Total & \multicolumn{2}{r}{\textbf{2.37M}} \\
\bottomrule
\end{tabular}
\caption{\label{tab:chart_to_table_datasets}
Pre-training datasets for learning chart representations include examples from numerous tasks that have paired chart images with table representations.
}
\end{table}

The existing table representation is used as is from the datasets, or, as described earlier, for a small fraction, tables are created programmatically. Tables are also normalized to a standardized format.

\subsection{Fine-tuning: Multi-task Loss}

After the pre-training stage which enables the ViT backbone to work better with charts, we use the synthetic data to fine-tune the model for the downstream task. We investigate two ways of incorporating the rationales available in the extended dataset.

The first one is by changing the task target from \textit{answer} to \textit{rationale, answer}. This has been shown to be effective in \cite{magister2023teaching}. We refer to this approach as \textbf{single-task setup}. However, it requires increased inference time by predicting the rationale, together with increased sequence length during training. The unintended side effect of training to predict jointly rationales and answers is that rationale tokens become equally important as the answer tokens.

The second one is inspired by \citet{hsieh2023distilling} which addresses both concerns by constructing a \textbf{multi-task setup} where the answer and rationale are treated as independent tasks. This can be done using different prefixes similar to T5 \cite{raffel2023exploring}, such as  \textit{"Rationale:"} and \textit{"Question:"}. The training loss balances the strength between the two tasks using a hyper-parameter $\mathbf{\lambda}$:

\begin{center}
    $\mathbf{Loss = (1 - \lambda) Loss_{ans} + \lambda Loss_{rat}}$
\end{center}

Our experiments are the first application of this setup for a multimodal task. We further confirm the observation from text domains that not only inference time remains constant, but quality also improves.

\section{Experiments} \label{sec:experiments}

We describe the general learning hyper-parameters for the pre-training and fine-tuning stages, followed by interpretation of the results.

\subsection{Setup}\label{sec:hyperparams}

\paragraph{Pre-training} We continue pre-training the PaLI-3 model with ViT unfrozen on the Chart2Table data mixture for \texttt{train\_steps=6K}, \texttt{batch\_size=256} with \texttt{learning\_rate=5e-3} with normalized square root decay using \texttt{decay\_factor=2e-6} and \texttt{dropout\_rate=0.1}.

\paragraph{Fine-tuning} We then freeze the ViT encoder and continue fine-tuning on the synthetic ChartQA dataset for \texttt{train\_steps=5K}, \texttt{batch\_size=256} with \texttt{learning\_rate=1e-3} with linear decay using \texttt{decay\_factor=1e-4} using \texttt{dropout\_rate=0.1}.

\paragraph{Multitask} We use $\mathbf{\lambda=0.5}$ and we do not find significant differences when using other values.

\subsection{Results on ChartQA}\label{sec:results}

We validate the effectiveness of the different techniques by reporting the downstream task performance on the ChartQA test set. All following experiments are on PaLI-3.

\paragraph{Pre-training} Continuing the pre-training stage for the PaLI-3 model using the Chart2Table mixture enables learning a better general representation of the charts. We intuitively expect that this better representation enables the model to more accurately identify quantities on the images. Indeed, we confirm this first through the results reported in Table~\ref{tab:pali3_pretraining_strategy}. Later, as we scale the dataset size, we show that this continues to play an important role.

\begin{table}[H]
\centering

\resizebox{\columnwidth}{!}{%
\begin{tabular}{lccc}
\toprule
\multirow{2}{5cm}{\textbf{Pre-training Strategy}} & \multicolumn{3}{c}{ChartQA (RA\%)} \\
\cmidrule{2-4}
 & \textbf{Avg.} & \textbf{Hum.} & \textbf{Aug.} \\
\midrule
Original PT \cite{chen2023pali3} & 70.00 & - & - \\
Chart2Table PT (our run) & \textbf{70.84} & 48.96 & 92.72 \\
\bottomrule
\end{tabular}%
}
\caption{\label{tab:pali3_pretraining_strategy}
PaLI-3 performance on ChartQA slightly increases with our chart-to-table pre-training phase.
}
\end{table}

As expected, the increase is predominantly in the augmented set, given that the pre-training mixture is constructed synthetically as well.

\paragraph{Singletask vs. Multitask} We first study the effect of introducing rationales only using the \textit{ChartQA-Rationale-S}. This only adds rationales to the original ChartQA dataset.

When using the rationales in singletask setup the performance difference is not significant compared to not using them. However, when used in the multitask setup, we note a quality improvement, particularly noticeable in the more difficult human-set. We refer to the former as \textit{Singletask-Rationale} and to the latter as \textit{Multitask-Rationale}  in Table~\ref{tab:pali3_single_vs_multi}.

\begin{table}[H]
\centering

\resizebox{\columnwidth}{!}{%
\begin{tabular}{lccc}
\toprule
\multirow{2}{5cm}{\textbf{Fine-tuning setup}} & \multicolumn{3}{c}{ChartQA (RA\%)} \\
\cmidrule{2-4}
 & \textbf{Avg.} & \textbf{Hum.} & \textbf{Aug.} \\
\midrule
C2T PT + Singletask-Rationale & 70.80 & 49.36 & 92.24 \\
C2T PT + Multitask-Rationale & \textbf{71.72} & 50.72 & 92.72 \\
\bottomrule
\end{tabular}%
}
\caption{\label{tab:pali3_single_vs_multi}
Multitask performance stands out compared to Singletask on the more difficult human-written set.}
\end{table}

We hypothesize that the improvement comes from better use of the rationales, guiding the model to internally produce a form of reasoning before producing the final answer. This is done without paying the cost predicting the rationales tokens.

\paragraph{Learning with augmented dataset} We use the ChartQA-Synth dataset from Table~\ref{tab:reasoning_datasets} for studying the extent to which we can transfer reasoning capabilities from PaLM-2 to PaLI-3.

We perform an ablation experiment to understand the role of the extra questions, rationales and pre-training stage and report our results in Table~\ref{tab:extraqar_results}.

We denote experiments using the original pre-trained checkpoint as \textit{Orig PT} and on the further pre-trained checkpoint with chart-to-table translation as \textit{C2T}. We report a clear improvement, further strengthening our observation that internal representation plays an important role.

\begin{table}[H]
\centering

\resizebox{\columnwidth}{!}{%
\begin{tabular}{lccc}
\toprule
\multirow{2}{5cm}{\textbf{Fine-tuning Setup}} & \multicolumn{3}{c}{ChartQA (RA\%)} \\
\cmidrule{2-4}
 & \textbf{Avg.} & \textbf{Hum.} & \textbf{Aug.} \\
\midrule
Orig PT + Singletask-ExtraQAR & 72.43 & 53.20 & 91.67 \\
Orig PT + Multitask-ExtraQAR & 73.15 & 55.20 & 91.10 \\
\midrule
C2T PT + ExtraQA \small{(w/o Rationale)} & 74.67 & 56.39 & 92.96 \\
\midrule
C2T PT + Singletask-ExtraQAR & 75.16 & 55.84 & \textbf{94.48} \\
C2T PT + Multitask-ExtraQAR  & 75.36 & 56.80 & 93.92 \\
\midrule
C2T PT + Singletask-ChartQA-Synth & 76.60 & 59.04 & 94.16 \\
C2T PT + Multitask-ChartQA-Synth & \textbf{77.28} & \textbf{60.88} & 93.68 \\
\bottomrule
\end{tabular}%
}
\caption{
Ablation results confirm the importance of each step in our recipe. \textit{ChartQA-Synth} is the mixture described in Table~\ref{tab:reasoning_datasets}}
\label{tab:extraqar_results}
\end{table}

We ran an experiment without rationales, but with the entire synthetically generated QA pairs. We note that the increase in examples ends up improving over the original ChartQA performance reported in Table~\ref{tab:pali3_pretraining_strategy}. However, the use of rationales continues to improve quality for both singletask and multitask setups. We observe that in high-data regimes, there is no longer a significant difference between the two.

Given the neutral impact of the multi-task setup at inference time, paired with slightly improved performance on the human-written queries of ChartQA, multi-task is the preferred option in practice. Further, we refer to the best performing fine-tuned setup in Table~\ref{tab:extraqar_results} as \textbf{ChartPaLI-5B}.

\begin{table*}[t]
\centering
\small
\begin{tabular}{llc}
\toprule
\textbf{Fine-tuned VLMs} (up to 55B) & Source & \textbf{ChartQA (RA\%)} \\
\midrule
Fuyu-8B & our eval, \cite{fuyu-8b} & 42.1 \\
Pix2Struct-1.3B & \cite{lee2023pix2struct} & 58.6 \\
MatCha-300M & \cite{liu2023matcha} & 64.2 \\
UniChart-201M & \cite{masry2023unichart} &  66.2  \\
ChartLlama-13B & \cite{han2023chartllama} & 69.6 \\
PaLI-5B & \cite{chen2023pali3} & 70.0 \\
PaLI-55B & \cite{chen2023palix} & 70.9 \\
PaLI-55B (Soft Mixture of Low-rank Experts) & \cite{wu2023omnismola} & 73.8 \\
ChartPaLI-5B & \textbf{our work} & \textbf{77.3} \\
\midrule
\multicolumn{2}{l}{\textbf{Hybrid VLMs/LLMs} (undisclosed size)} \\
\midrule
GPT-4V [4-shot with CoT] & \cite{openai2023gpt4} & 78.5 \\
DePlot-300M + FlanPaLM + Codex with PoT SC & \cite{liu2023deplot} & 79.3 \\
Gemini Ultra [0-shot] & \cite{gemini} & 80.8 \\
ChartPaLI-5B + PaLM 2-S PoT SC @ 5 &  \textbf{our work} & \textbf{81.3} \\
\bottomrule
\end{tabular}
\caption{\label{tab:sota_over_baseline}
State-of-the-art performance among fine-tuned VLMs on ChartQA benchmark.
}
\end{table*}

\subsection{Results on FigureQA and PlotQA}\label{sec:addtl_results}

ChartQA is currently the most challenging benchmark. To prove that our method is general, we investigate performance on related chart understanding tasks, FigureQA \cite{} and PlotQA \cite{}. We study 3 operation regimes: (i) \textbf{zero-shot}: no task-specific pre-training or fine-tuning, (ii) \textbf{quick adaptation}: 1K fine-tuning steps and (iii) \textbf{convergence}: 5K fine-tuning steps. We report relaxed accuracy on 10K examples from validation set for FigureQA (ref. Table~\ref{tab:figureqa_results} and from test set from PlotQA (ref. Table~\ref{tab:plotqa_results}).

\begin{table}[H]
\centering

\resizebox{\columnwidth}{!}{%
\begin{tabular}{lccc}
\toprule
\multirow{2}{3cm}{\textbf{Model}} & \multicolumn{3}{c}{FigureQA RA\% (v1 | v2)} \\
\cmidrule{2-4}
 & \textbf{ZShot} & \textbf{Quick} & \textbf{Conv} \\
\midrule
PaLI-3 (original) & 41.9 | 42.4 & 57.2 | 58.1 & 89.9 | 89.3 \\
ChartPaLI-5B & \textbf{51.0 | 51.2} & \textbf{92.7 | 93.0} & \textbf{96.3 | 96.2} \\
\bottomrule
\end{tabular}%
}
\caption{\label{tab:figureqa_results} ChartPaLI-3 exhibits strong generalization on FigureQA task, for which no examples are present in pre-training or fine-tuning}
\end{table}

For PlotQA, images from the training subset are present in our pre-training mixture, while validation and test subset images are not. Therefore, we do not study zero-shot performance, as training images would give an unfair advantage.

\begin{table}[H]
\centering

\resizebox{\columnwidth}{!}{%
\begin{tabular}{lccc}
\toprule
\multirow{2}{3cm}{\textbf{Model}} & \multicolumn{2}{c}{PlotQA RA\% (v1 | v2)} \\
\cmidrule{2-3}
 & \textbf{Quick adapt.} & \textbf{Convergence} \\
\midrule
PaLI-3 (original) & 62.0 | 15.7 & 71.5 | 23.6 \\
ChartPaLI-5B & \textbf{79.1 | 53.3} & \textbf{86.0 | 70.7} \\
\bottomrule
\end{tabular}%
}
\caption{\label{tab:plotqa_results} ChartPaLI-3 exhibits strong generalization on FigureQA task, for which no examples are present in pre-training or fine-tuning}
\end{table}

ChartPaLI-5B outperforms PaLI-3 in all operation regimes. In general, our recipe significantly increases chart understanding performance when running only a few quick adaptation steps.

In particular we report SoTA performance regime for FigureQA (roughly 96\%+) and the very strong relative performance on the difficult PlotQA v2 (roughly +47.1\% at convergence time).

\subsection{Errors and Challenges}\label{sec:errors}

To understand the effect of our method and investigate further opportunities for improvements, we manually looked at predictions on the ChartQA validation set. We compared baseline PaLI-3 model outputs with the model fine-tuned with our recipe and share our observations below. We report our findings below.

\paragraph{General} The model predicts the rationale\footnote{Although the table is not used during inference, the rationales contain the word \textit{table} due to its use in prompts.} or the answer, depending on the task prefix. Because the answer is not conditioned on the rationale, it can differ. One general improvement area we note is the ability to extract necessary intermediate quantities (Fig.~\ref{fig:correct_intermediates_and_answer}) and operate with them (Fig.~\ref{fig:correct_intermediates_and_answer_2}).

\begin{figure}[ht]
\center
\includegraphics[width=\columnwidth]{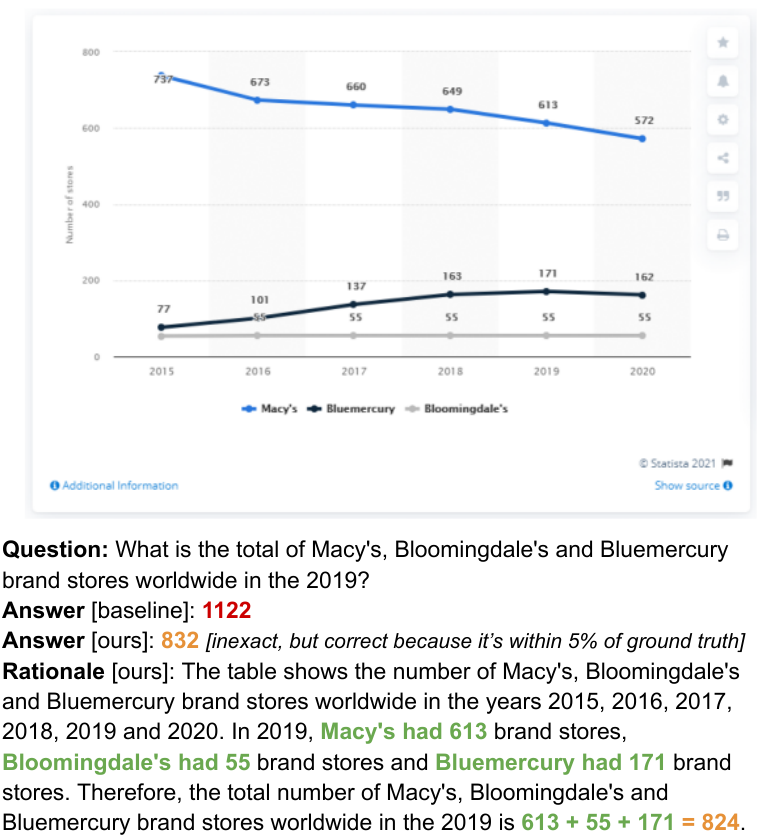}
\caption{Correct numeric approximations on answers.}
\label{fig:almostcorrect_numerics}
\end{figure}

\paragraph{Numerical reasoning} Despite improvements, computation of mathematical expressions continues to be very challenging. The rationales correctly extract (Fig.~\ref{fig:almostcorrect_numerics}) or infer chart values when missing (Fig.~\ref{fig:infer_chart_values}), however the computed value is frequently incorrect. This does not always prevent the final answer to be correct (Fig.~\ref{fig:incorrect_arithmetic_but_correct_answer}). This seems in line with observations by \citet{wang-etal-2023-towards}, who also conclude that corruption of the chain-of-thought reasoning trace does not always degrade the final answer. Due to the frequency of this numeric computation error, we explore a simple refining technique in Section~\ref{sec:pyrefine}.

\paragraph{Color reasoning} Our synthetic data does not have color metadata, as only the table was used in the generation process. Therefore the model continues to struggle when the reasoning trace requies working with colors (Fig.~\ref{fig:error_color_reasoning}). Thus, this is an area worth of investigating next and has applicability well beyond the specifics of chart understanding.

\paragraph{Complex reasoning} Reasoning about multiple values and checking for a matching condition which requires arithmetic computations is another example of a remaining difficult task (Fig.\ref{fig:complex_reasoning}, Fig.\ref{fig:complex_reasoning_2}). The increased complexity stemming from internal inability of VLMs to perform numeric operations paired with enumerating chart elements through semantic descriptions is likely fairly difficult to achieve without the use of external tools.

\paragraph{Task leakage} Due to the training methodology, we observe that when conditioned with the \textit{Question} task prefix, the model may behave similarly as to when \textit{Rationale} prefix is used. Sometimes, instead of directly outputting an answer, the model may generate a longer explanation that resembles a rationale or a fragment of rationale.

\subsection{Refinement with Program of Thoughts} \label{sec:pyrefine}

Despite the improved ability to construct numeric equations using the required values on the charts (Fig.~\ref{fig:almostcorrect_numerics}), the exact numeric computation continues to be wrong. This is unsurprising, since both the visual and the language backbone treat numbers as tokens. Making the problem worse, the character sequence forming a number may be split and encoded in arbitrary chunks. \citet{chen2023program} have proposed replacing chain-of-thoughts (CoT) prompting with program-of-thoughts (PoT) to enable delegation of the arithmetic computation to a program interpreter. This has previously been explored by \citet{liu2023deplot}, however in a much more computationally involved setup than the one we describe further.

Through our fine-tuning approach, both singletask and multitask setups can be used produce CoT rationales for which an LLM prompted with PoT can write the equivalent code for performing the numeric computation.

We take the approach of using a simple 4-shot prompt (Fig.~\ref{fig:python_prompt}) constructed on the validation set to generate code using PaLM 2-S for performing the numeric computation that is present in a rationale. We run this online refinement, only if the rationale contains an arithmetic operator ('+', '-', '/' or '*').

Self-consistency is an effective way to improve chain-of-thoughts rationales by selecting an answer with majority voting from a pool of sampled rationales \cite{wang2023selfconsistency}. We apply this approach, by sampling with temperature $\tau_{Rat}=0.4$ and generate $N=5$ rationales that are then refined with PaLM 2-S using temperature $\tau_{Ref}=0.0$.

\begin{table}[H]
\centering

\resizebox{\columnwidth}{!}{%
\begin{tabular}{lccc}
\toprule
\multirow{2}{5cm}{\textbf{Setup}} & \multicolumn{3}{c}{ChartQA (RA\%)} \\
\cmidrule{2-4}
 & \textbf{Avg.} & \textbf{Hum.} & \textbf{Aug.} \\
\midrule
ChartPaLI-5B (from Table~\ref{tab:extraqar_results}) & 77.28 & 60.88 & 93.68 \\
ChartPaLI-5B + PaLM 2-S PoT & 80.80 & 67.92 & 93.68 \\
ChartPaLI-5B + PaLM 2-S PoT SC @ 5 & \textbf{81.32} & \textbf{68.96} & 93.68 \\
\bottomrule
\end{tabular}%
}
\caption{\label{tab:pot_refinement}
PoT refinement improves performance on the human set, while not affecting the augmented set.}
\end{table}

The results presented in Table~\ref{tab:pot_refinement} highlight the utility of the method, particularly with K=5 for self-consistency. They also highlight the simplicity of the augmented set compared to the human set, for which the refinement does not have an impact. Either the augmented set contains no arithmetic computations or they are simple enough for the fine-tuned VLM to already get right.

\section{Performance Overview}

We position our results relative to existing prior work in Table~\ref{tab:sota_over_baseline}. We extracted the results from the referenced papers, with the exception of the Fuyu-8B \cite{fuyu-8b} model. We performed our own evaluation as the authors have not provided the results on the ChartQA benchmark. 

Our work significantly outperforms prior models specialized on the ChartQA benchmark. Concurrent to our work, ChartLlama-13B also uses synthetic data generated, but with a fairly different approach. Although outside the scope of our work, it may be that the approach took to train the much smaller MatCha and UniChart models may be combinable with the approach we presented in this work, leading to possible improved performance with even less computational resources.

The method introduced in this work can be uniquely combined with much larger models through rationale generation. As shown in the results, rationales generated by VLMs can suffice for larger LLMs to effectively operate on, providing a text-representation of the chart conditioned on the question. Our method matches the recently introduced Gemini Ultra model and outperforms previous approaches.

\section{Future Work}\label{sec:futurework}

We highlighted several drawbacks of our approach in Section~\ref{sec:errors}. The training mixtures do not have examples where colors are used to construct reasoning examples. Bootstrapping such examples, for example by running a smaller sized model with questions that extract color related information, then combines them, would likely improve quality. Very complex reasoning examples are also limited. Specifically,  semantically identifying chart elements and performing numeric computations to solve questions would further improve quality.

\section{Conclusion}\label{sec:conclusion}

We introduced a novel recipe that significantly improves the reasoning capabilities of VLMs. Applied to PaLI-3, our method significantly outperforms even the 10x larger PaLI-X on the ChartQA benchmark, establishing a new state-of-the-art. We demonstrate how the pre-training stage improves downstream performance. Our synthetic data generation technique coupled with the use of a multi-task setup, successfully transfers reasoning capabilities from larger LLMs to smaller VLMs. Moreover, our method enables a computationally more expensive setup where predicted rationales are refined using program-of-thoughts with PaLM 2-S. The composite solution outperforms Gemini Ultra and GPT-4V on the ChartQA benchmark.

\section{Limitations}\label{sec:limitations}

We acknowledge limitations of our approach. 

\paragraph{Table representation} Although our final model works on pixels only, our synthetic data generation method requires having access to a table version of the charts for leveraging LLMs to construct rationales, additional question/answer pairs, etc for the training datasets. Although it is likely that inferred tables or output of an OCR model may replace to some degree the presence of gold tables, it will likely affect final model quality.

\paragraph{PaLI-3} The pre-training and fine-tuning recipe for synthetic data creation, as well as the training methodology should be applicable broadly on open source models as well. However, we acknowledge that the choice of PaLI-3, a proprietary flavor of VLMs, is not as a good of a choice as an open source flavor available externally.

\paragraph{Risks associated with synthetic dataset} Since the method for constructing our dataset relies on LLMs, there are certain inherent risks that come with that, for example that of hallucination. Although our technique extends the publicly available ChartQA dataset, additional care needs to be taken into account when planning to apply it for releasing models or dataset openly. Although the metrics are state-of-the-art, it cannot be guaranteed that model outputs can't be abused if trained in this manner.

\paragraph{Reasoning limitations} We acknowledge limitations stemming from the empirical prompt creation process, which is based on human inspection of model errors. LLM capabilities used for the synthetic data creation, although impressive, continue to have numerous limitations as reported by the community.

\section{Acknowledgements}

We thank Srinivas Sunkara and Maria Wang for their contributions on the infrastructure that enabled us to run these experiments. Further, we thank Xi Chen for his tireless support and insights into PaLI-3 details and training recipes and Cheng-Yu Hsieh and Yasuhisa Fujii for the detailed discussions on the multi-task setup. Daniel Keysers and Radu Soricut have provided detailed feedbacks that significantly improved the paper.  Matt Sharifi and Ewa Dominowska provided senior leadership support for this work.

Lastly, feedback from anonymous reviewers rPKR and 453J motivated running additional experiments further strengthening the contribution of this work by showcasing the method is generally applicable.

\newpage

\bibliography{acl2024}
\bibliographystyle{acl_natbib}

\clearpage
\appendix

\section{Prompts for PaLM-2}

\begin{figure}
\includegraphics[width=0.4\textwidth]{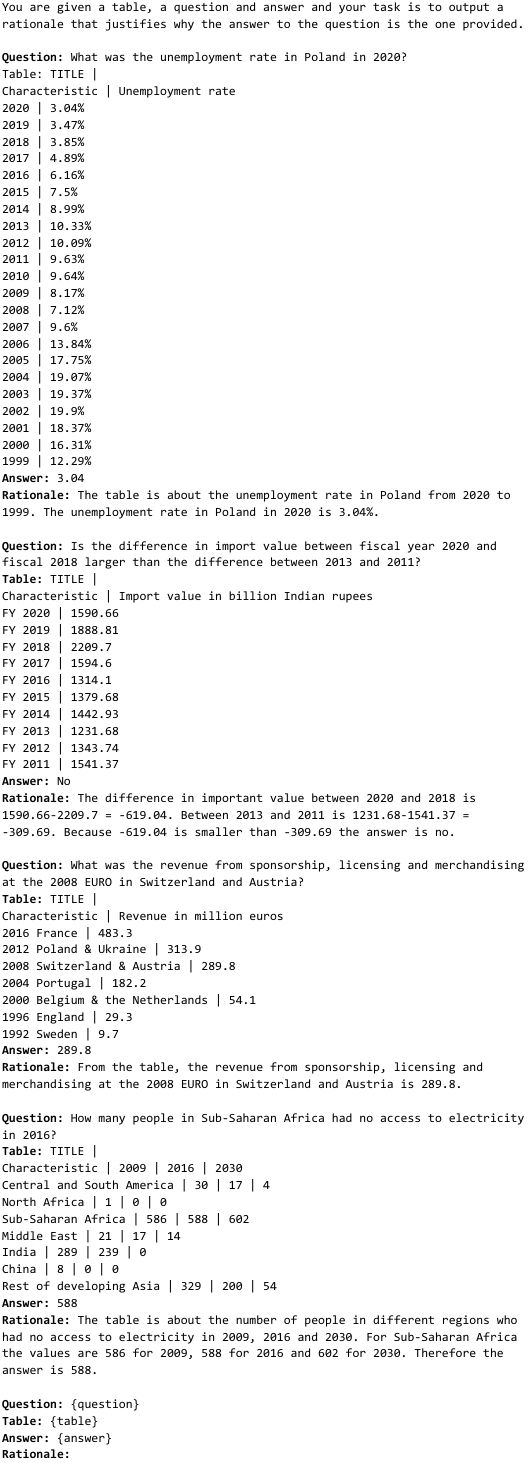}
\caption{The input template, with a 4-shot prompt, for generating the ChartQA-Rationale-S dataset using PaLM 2-S.}
\label{fig:rationale_prompt}
\end{figure}

\begin{figure}
\includegraphics[width=0.4\textwidth]{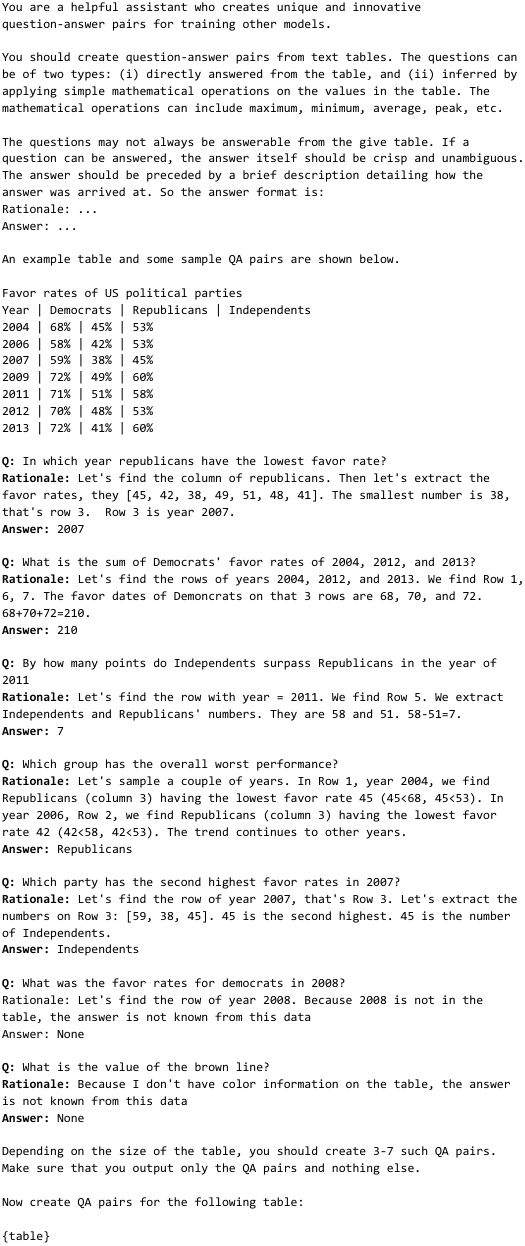}
\caption{The input template, with a 1-shot prompt, for generating the ChartQA-ExtraQAR-S/L datasets using PaLM 2-S/L.}
\label{fig:extraqa_prompt}
\end{figure}

We use PaLM 2-S and PaLM 2-L throughout this work. Here we describe the prompts used for the different purposes. Our \textit{ChartQA-Rationale-S} dataset is a straightforward augmentation of the ChartQA dataset, by predicting the rationales using the table, answer and question. For this, we have constructed the prompt illustrated in Figure~\ref{fig:rationale_prompt}. The \textit{ChartQA-ExtraQAR-S/L} datasets are constructed using PaLM 2-S/L respectively for which we extended the 1-shot prompt provided by \cite{liu2023deplot}. We chose this prompt for simplicity and for it already containing several diverse question examples. The prompt is illustrated in Figure~\ref{fig:extraqa_prompt}.

\begin{figure}
\includegraphics[width=0.4\textwidth]{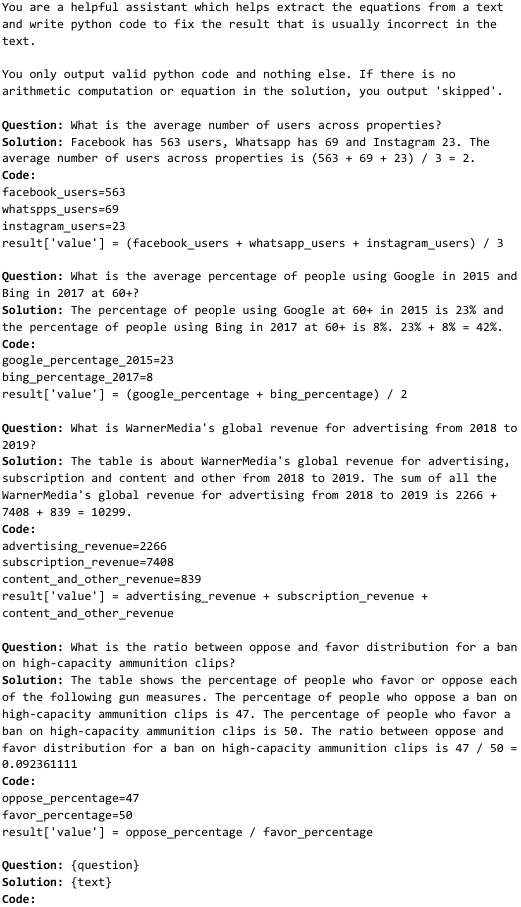}
\caption{The input template, with a 4-shot prompt, for refining arithmetic operations with python code using PaLM 2-S.}
\label{fig:python_prompt}
\end{figure}

Lastly, we describe an online refinement of the rationale prediction using program-of-thoughts in Section~\ref{sec:pyrefine}. For this, we manually constructed the prompt illustrated in Figure~\ref{fig:python_prompt}. This was built by inspecting a few validation errors when the numeric values computed by the VLM were wrong.

\section{Generated Examples}

\begin{figure}
\includegraphics[width=0.5\textwidth]{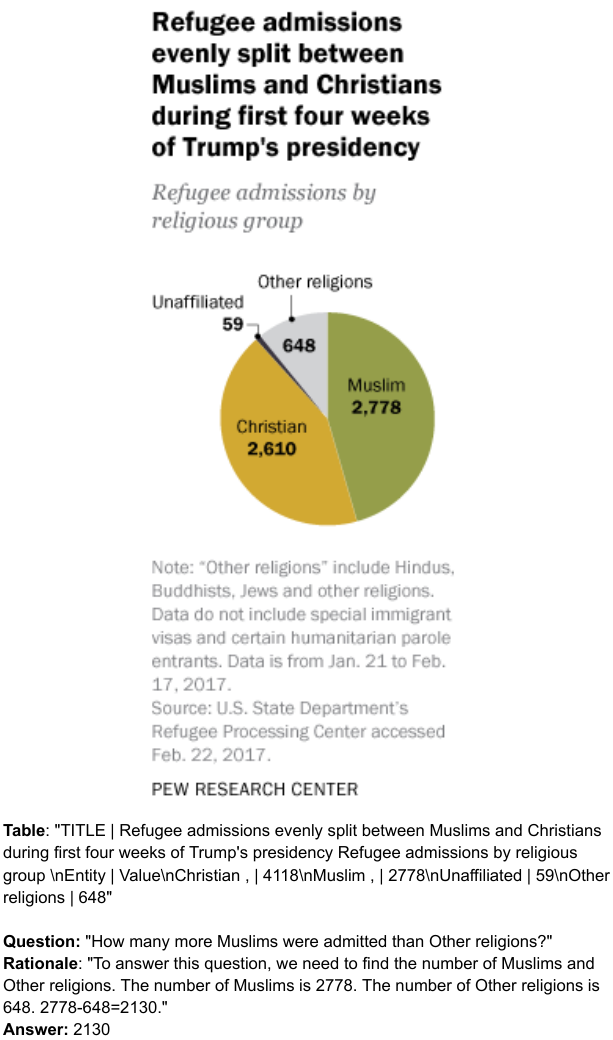}
\caption{\textit{ChartQA-ExtraQAR-S/L}: Example of synthesized (Question, Answer, Rationale) pair with PaLM-2 using the table}
\label{fig:extraqar_example}
\end{figure}

\paragraph{Licensing} As we redistribute certain data artifacts, we note that the ChartQA dataset at the time of this writing is marked as GPL v3.0 \footnote{https://github.com/vis-nlp/ChartQA/blob/main/LICENSE}. In this section we provide visual examples of our synthetically generated training datasets, using PaLM 2-S/L models, as well as the programmatically generated templates for mathematical computations. Figure~\ref{fig:extraqar_example} contains an example of synthesized example using only the table representation. The question, answer and rationale cover an aspect of the table and are generated together with 3-5 other questions.

\begin{figure}
\includegraphics[width=0.5\textwidth]{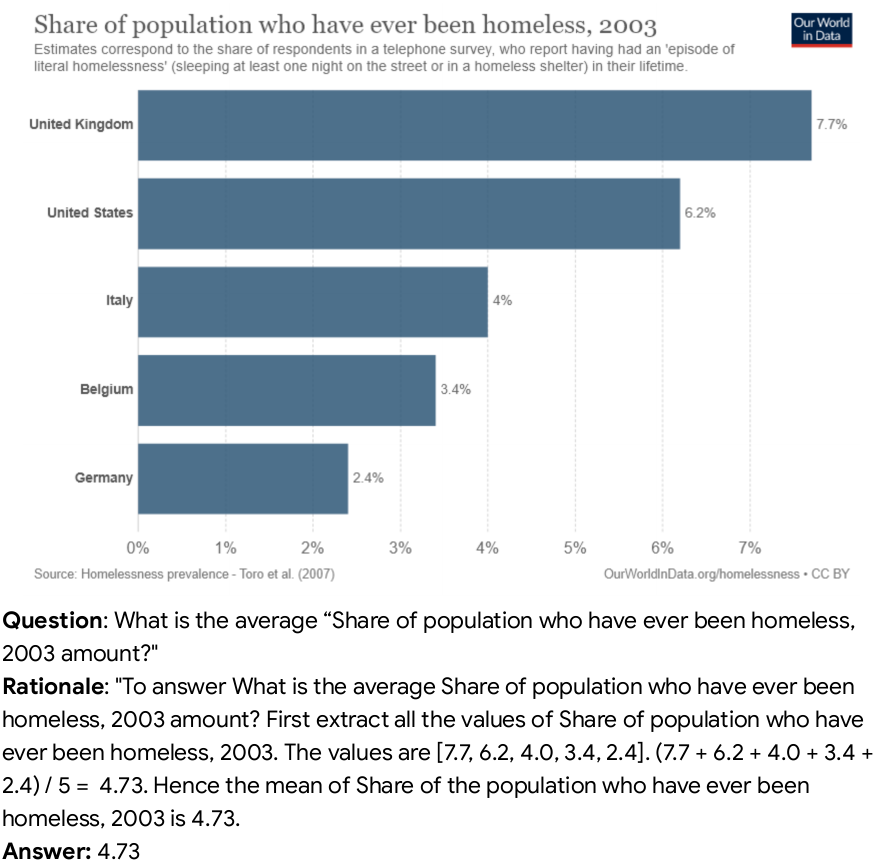}
\caption{\textit{ChartQA-ArithmQAR}: Example of programmatically generated (Question, Answer, Rationale) pair}
\label{fig:arithm_example_1}
\end{figure}

\begin{figure}
\includegraphics[width=0.5\textwidth]{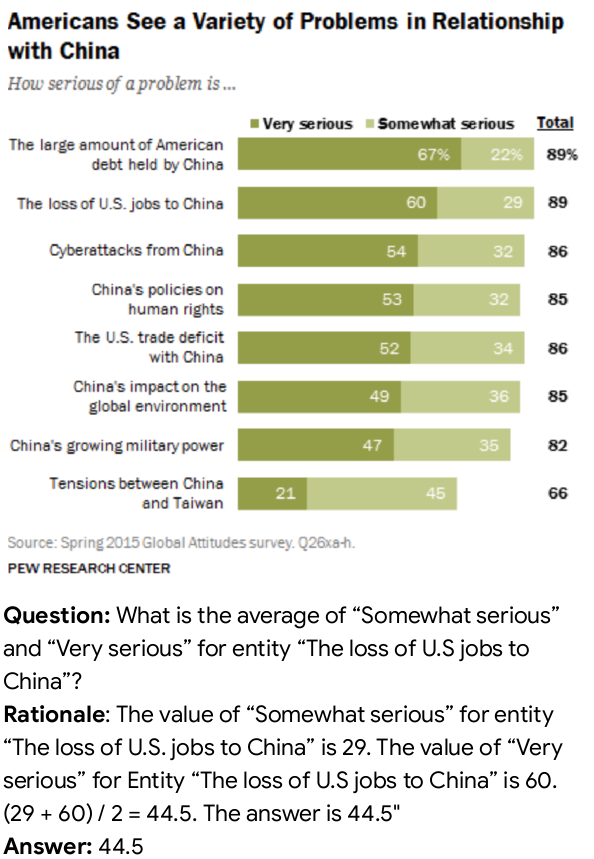}
\caption{\textit{ChartQA-ArithmQAR}: Example of programmatically generated (Question, Answer, Rationale) pair}
\label{fig:arithm_example_2}
\end{figure}

Figure~\ref{fig:arithm_example_1} and Figure~\ref{fig:arithm_example_2} are examples of a programmatically generated questions based on the template to compute the mean. The markdown table provided as input is processed through a function that takes the corresponding values and outputs all the elements, including the reasoning trace in the rationale for computing the mean as shown in the figure. 

\section{Model Outputs}

\begin{figure}
\centering
\includegraphics[width=0.4\textwidth,clip]{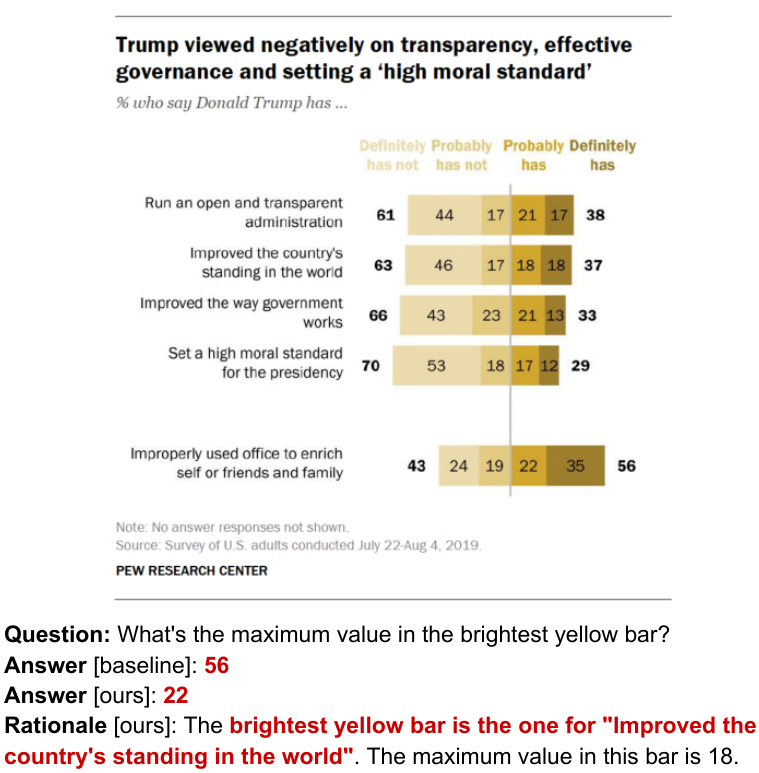}
\caption{Matching the colors with content is weak.}
\label{fig:error_color_reasoning}
\end{figure}

\begin{figure}
\centering
\includegraphics[width=0.4\textwidth,clip]{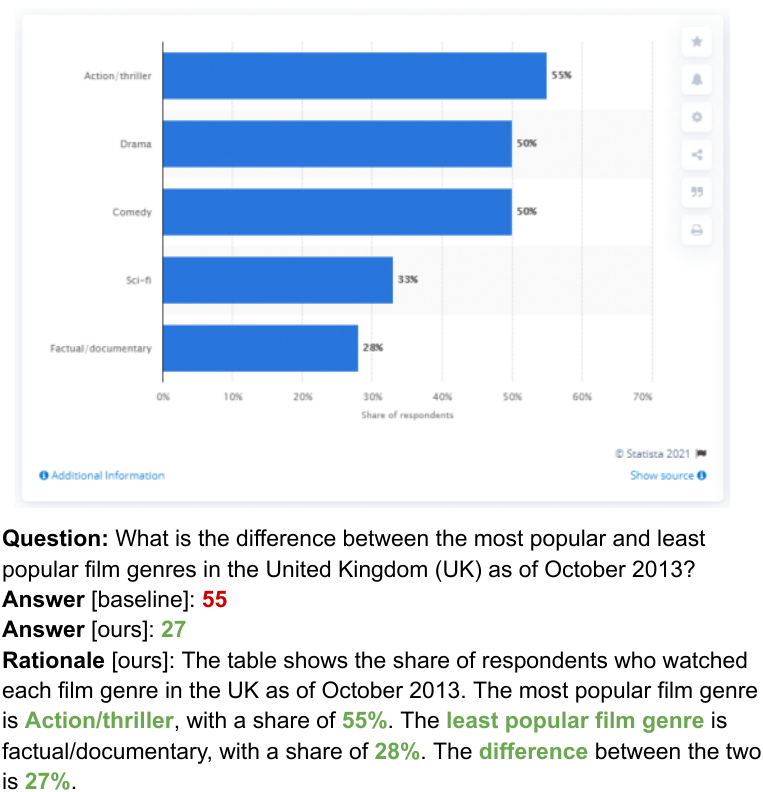}
\caption{Excellent extraction of intermediate values.}
\label{fig:correct_intermediates_and_answer}
\end{figure}

\begin{figure}
\centering
\includegraphics[width=0.4\textwidth,clip]{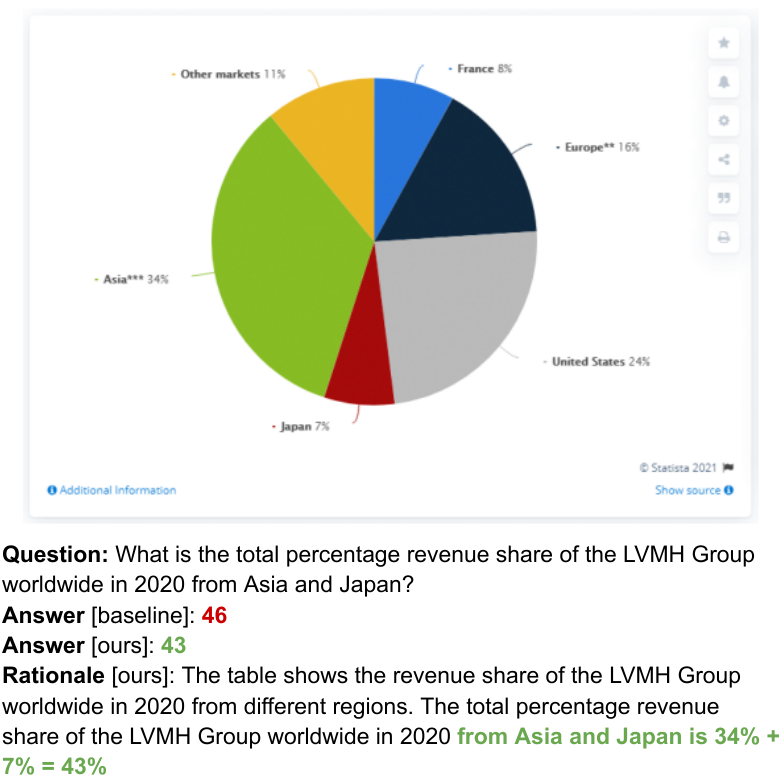}
\caption{Correct handling of extracted quantities.}
\label{fig:correct_intermediates_and_answer_2}
\end{figure}

\begin{figure}
\centering
\includegraphics[width=0.4\textwidth,clip]{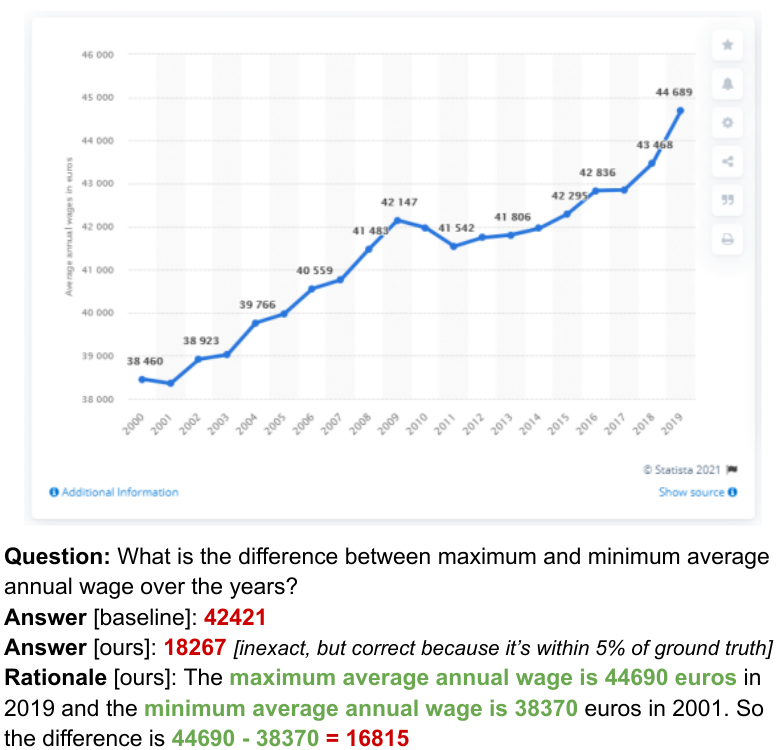}
\caption{Strong ability to infer missing values.}
\label{fig:infer_chart_values}
\end{figure}

\begin{figure}
\centering
\includegraphics[width=0.4\textwidth,clip]{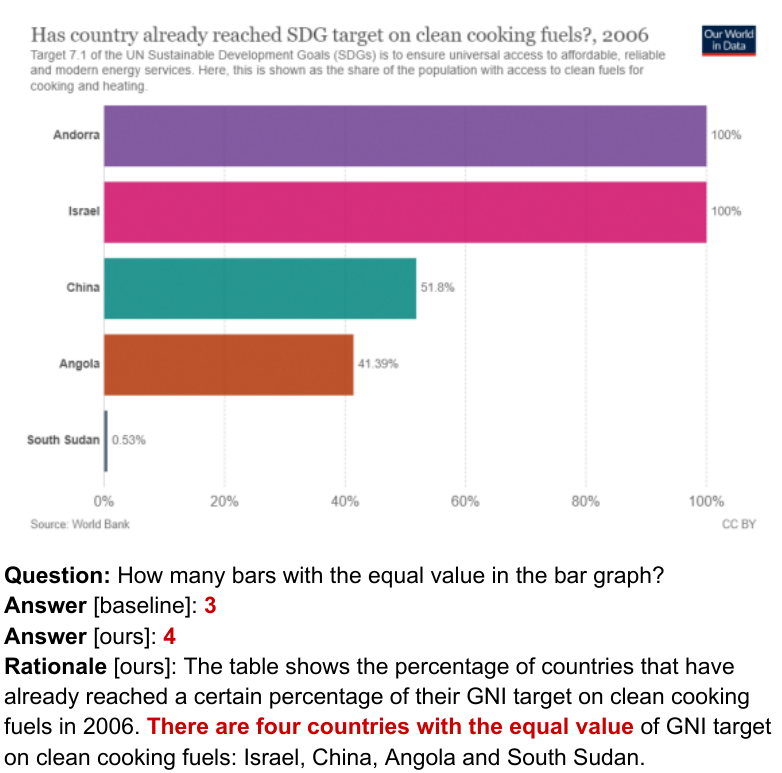}
\caption{Checking for equality among multiple values requires even better reasoning abilities.}
\label{fig:complex_reasoning}
\end{figure}

\begin{figure}
\centering
\includegraphics[width=0.4\textwidth,clip]{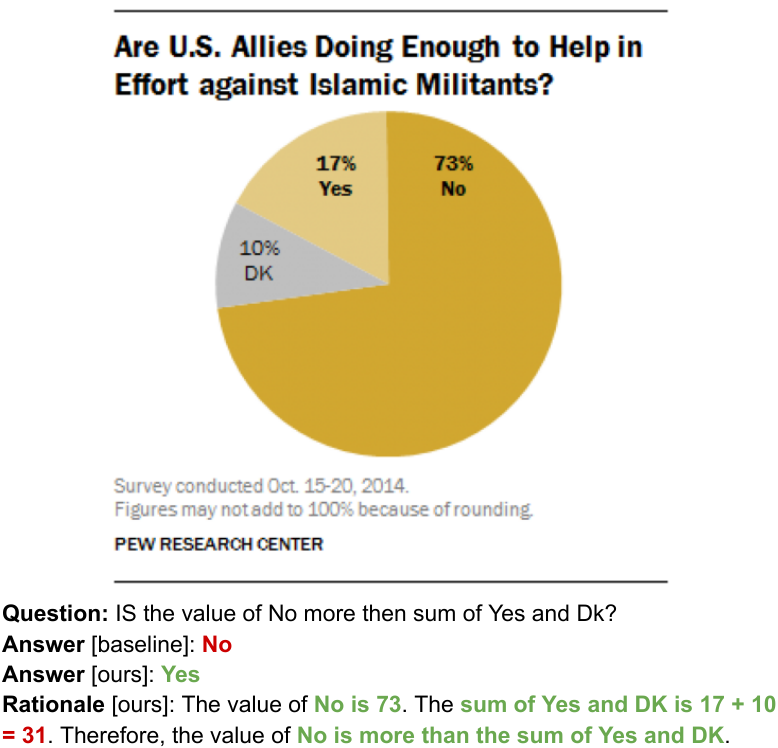}
\caption{Despite incorrect arithmetic results, final answer can still be correct}
\label{fig:incorrect_arithmetic_but_correct_answer}
\end{figure}

\begin{figure}
\centering
\includegraphics[width=0.4\textwidth,clip]{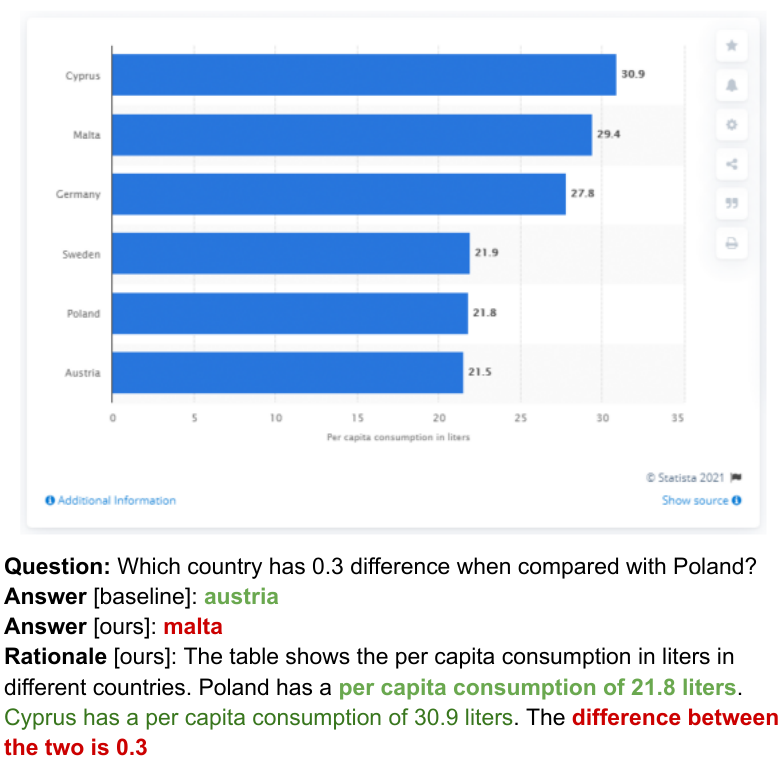}
\caption{Both answer and rationale can be wrong when it comes to enumerating values and checking more complex numerical conditions.}
\label{fig:complex_reasoning_2}
\end{figure}

In this section we provide examples that accompany our analysis of the model behavior. We highlighted impressive performance, as can be seen in  Figure~\ref{fig:correct_intermediates_and_answer},  Figure~\ref{fig:correct_intermediates_and_answer_2} and Figure~\ref{fig:infer_chart_values}. However, we noted several limitations as well, as can be seen in Figure~\ref{fig:error_color_reasoning}, Figure~\ref{fig:complex_reasoning} and Figure~\ref{fig:complex_reasoning_2}.

\clearpage
\section{Synthetic Dataset: Statistics}
\label{sec:statistics_dataset}

We report the final dataset distribution for our synthetically generated examples. We follow the type of descriptions reported in the paper introducing the benchmark \cite{masry2022chartqa}. In Table~\ref{tab:reasoning_datasets} we describe the type of question generated. We note that we cannot generate, due to the use of a text-based language model, Visual and Compositional and Visual questions, so we only have Data Retrieval and Compositional. The use of visual captions may enable generating these other types.

We report in Table~\ref{tab:chart_types_chartqal} and Table~\ref{tab:chart_types_chartqas} the number of examples (question, answer, rationale) generated for each type of chart and source. We note that the total number of chart images does not change and they are the original ones from ChartQA.

\begin{table}[h]
\centering
\begin{tabular}{lrrrrrr}
\toprule
Source/Graph Type & Pew & Statista-Hum & OWID & OECD & Statista-Aug & Total \\
\midrule
H-Bar             & 2390 & 1140 & 1270 & -    & 13811            & 18611  \\
V-Bar             & -    & 4999 & -    & 358  & 31101            & 36458 \\
Pie               & 1162 & 1416 & 4    & -    & 366              & 2948  \\
Line              & 1101 & 1615 & 663  & 270  & 5190             & 8839  \\
\bottomrule
\end{tabular}
\caption{
\label{tab:chart_types_chartqal}
\resizebox{0.7\textwidth}{!}{\small Frequency of examples by chart types and sources for  \textbf{ChartQA-ExtraQAR-L}.}}
\end{table}

\begin{table}[h]
\centering
\begin{tabular}{lrrrrrr}
\toprule
Source/Graph Type & Pew & Statista-Hum & OWID & OECD & Statista-Aug & Total \\
\midrule
H-Bar             & 3561 & 1594 & 1923 & -    & 18777            & 25855  \\
V-Bar             & -    & 7061 & -    & -    & 42776            & 49837 \\
Pie               & 1525 & 1866 & 6    & 504    & 366             & 4375  \\
Line              & 1468 & 2268 & 1075  & 410  & 7406             & 12627  \\
\bottomrule
\end{tabular}
\caption{
\label{tab:chart_types_chartqas}
\resizebox{0.7\textwidth}{!}{\small Frequency of examples by chart types and sources for  \textbf{ChartQA-ExtraQAR-S}.}}
\end{table}

\end{document}